\documentclass[letterpaper, 10 pt, conference]{support/ieeeconf}
\usepackage{times}
\usepackage[pdftex]{graphicx}
\usepackage{subfigure}
\usepackage{amsmath,amssymb,amsopn,amstext,amsfonts}
\usepackage{cancel}
\usepackage[space]{cite}
\usepackage{pdfsync}
\usepackage{balance}
\usepackage{color}
\usepackage{mathtools}
\usepackage[ruled,vlined]{algorithm2e}
\usepackage{bm}

\usepackage{diagbox}
\usepackage{float}
\usepackage{epstopdf}
\usepackage{pifont}
\usepackage{fixltx2e}
\usepackage{amsmath}
\usepackage{multirow}
\usepackage{booktabs}
\usepackage{url}
\usepackage{svg}
\usepackage{bm}
\usepackage{threeparttable}
\usepackage[linkcolor=black,citecolor=black,urlcolor=black,colorlinks=true]{hyperref}

\newcommand{\tp}{^{\mathrm{T}}}

\graphicspath{{./figures/}}
\DeclareGraphicsExtensions{.png,.jpg,.eps,.pdf}
\IEEEoverridecommandlockouts
\makeatletter
\def\endthebibliography{%
	\def\@noitemerr{\@latex@warning{Empty `thebibliography' environment}}%
	\endlist
}

\title{\textbf{A Linear and Exact Algorithm for Whole-Body Collision Evaluation via Scale Optimization}}
\author{
        Qianhao Wang $^{\dag}$, 
        Zhepei Wang $^{\dag}$, 
        Liuao Pei, 
        Chao Xu,
        and Fei Gao
	    \thanks{\textbf{${\dag}$ Equal contribution.}}        
         \thanks{All authors are with the College of Control Science and Engineering, Zhejiang University, Hangzhou, 310027, China, and also with the Huzhou Institute of Zhejiang University, Huzhou, 313000, China. Email:{\tt\small \{qhwangaa, wangzhepei, fgaoaa\}@zju.edu.cn}}} 

\begin{document}

\maketitle
\thispagestyle{empty}
\pagestyle{empty}

\begin{abstract}
Collision evaluation is of essential importance in various applications.
However, existing methods are either cumbersome to calculate or not exact.
Therefore, considering the cost of implementation, most whole-body planning works, which require evaluating collision between robots and environments, struggle to tradeoff between accuracy and computationally efficiency.
In this paper, we propose a zero-gap whole-body collision evaluation that can be formulated as a low-dimensional linear programming.
This evaluation can be solved analytically in linear complexity.
Moreover, the method provides gradient efficiently, making it accessible to optimization-based applications.
Additionally, this method provides support for 
obstacles represented by either points or hyperplanes.
Experiments on the widely used aerial and car-like robots validate the versatility and practicality of our method.
\end{abstract}

\section{Introduction}
\label{sec:Introduction}
\begin{figure}[!t]
	\centering
    \vspace{0.2cm}
	\includegraphics[width=1\linewidth]{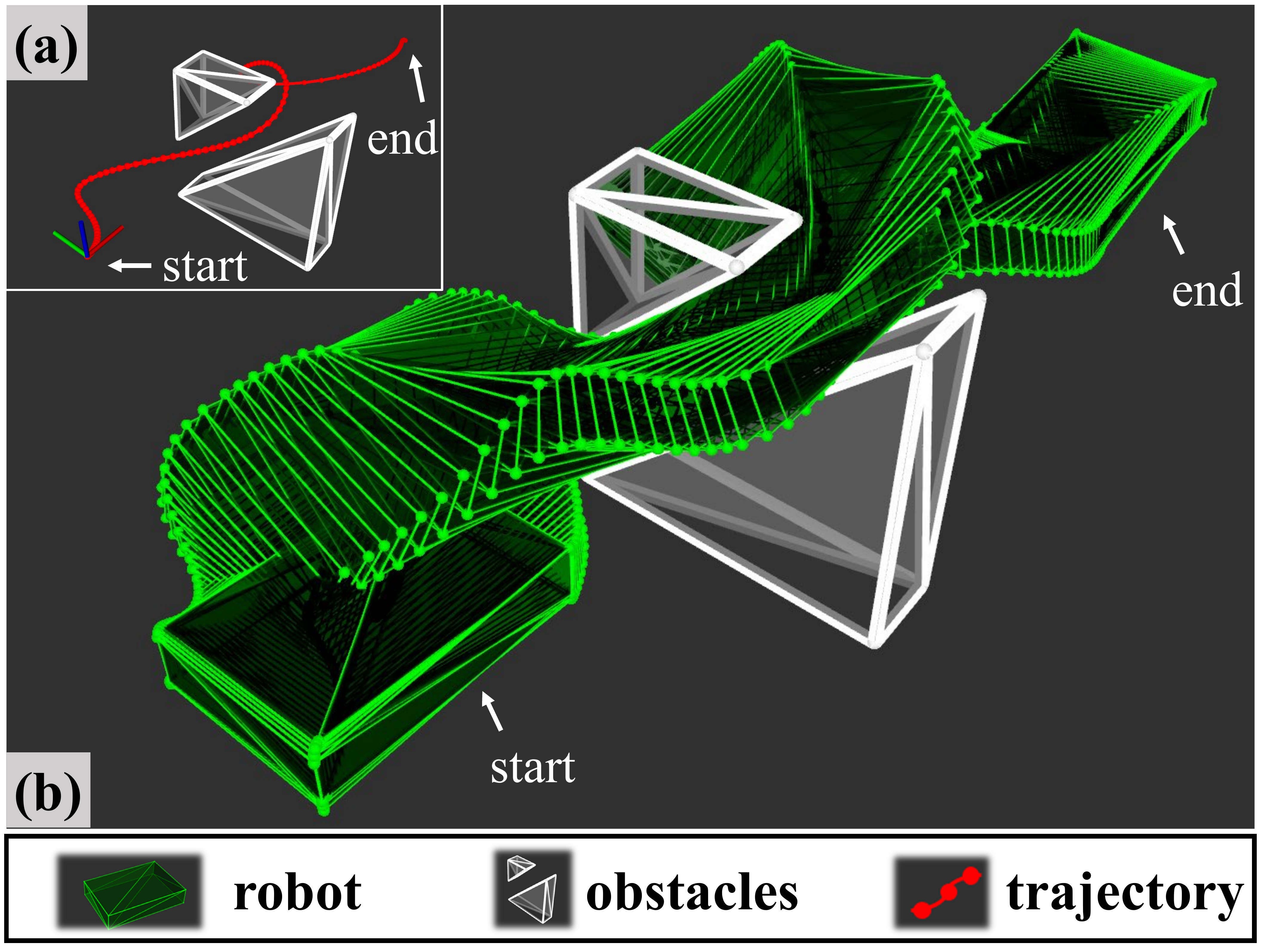}
	\caption{
        The application of a whole-body SE(3) planning problem which requires the multicopter to fly from the start to the end. 
        The green rectangles are the snapshots of the multicopter which flies along the red trajectory to cross the narrow gap built by the two white obstacles.
	}
	\label{fig:toutu}
    \vspace{-1cm}
\end{figure}

Collision Evaluation is critical in a variety of fields, such as physics engines, computer graphics and robot navigation.
In recent years, with the development of autonomy, an increasingly large number of robots are deployed in complex real-world scenarios, where they may be requested to navigate through dense and highly dynamic environments, as illustrated in Fig.\ref{fig:car}, or even to cross narrow gaps of similar size to themselves as shown in Fig.\ref{fig:toutu}.
These planning problems where the robot's shape has to be taken into account, namely whole-body planning, urgently require exact and efficient collision evaluation of robots and environments.

Generally, there are several expectations for a collision evaluation method.
\textbf{(1)} Exactitude: no gap or approximation with the true value is expected.
\textbf{(2)} Efficiency: low computational overhead is required. 
\textbf{(3)} Locality: instead of a rigorous enumeration of each obstacle~\cite{gilbert1994new,tracy2022differentiable}, the method only focus on a small number of obstacles around the robot.

Vast different approaches \cite{gilbert1988fast,cameron1997enhancing,van2003collision,gilbert1994new,tracy2022differentiable} have been introduced in response to the above mentioned requirements.
However, when applied to whole-body robot motion planning, they may suffer from difficulty in gradient calculation and struggle to deal with different map representations.
In detail, there are some intractable challenges for collision evaluation in whole-body planning.
\textbf{(1)} The method should be capable of providing precise gradients efficiently, which allows it to be applied in optimization-based frameworks.
\textbf{(2)} With the progress of environment reconstruction and SLAM technology, various map representations such as point clouds~\cite{ji2021mapless}, surfaces~\cite{wang2022geometrically}, grid maps~\cite{zhou2020egoplanner} and semantic information~\cite{ryll2020semantic}, are active in different planning works.
The ability to deal with different map representations that may contain lots of redundant information without overly complex pre-processing is required eagerly.
\textbf{(3)} It is unacceptable to make the dimension of the problem increase dramatically for collision avoidance, such as Zhang's work~\cite{zhang2020optimization} that introduces numerous dual variables in trajectory optimization.
Because this can lead to the optimization becoming hard to solve.

\begin{figure}[t]
	\centering
    \vspace{0.2cm}
	\includegraphics[width=1\linewidth]{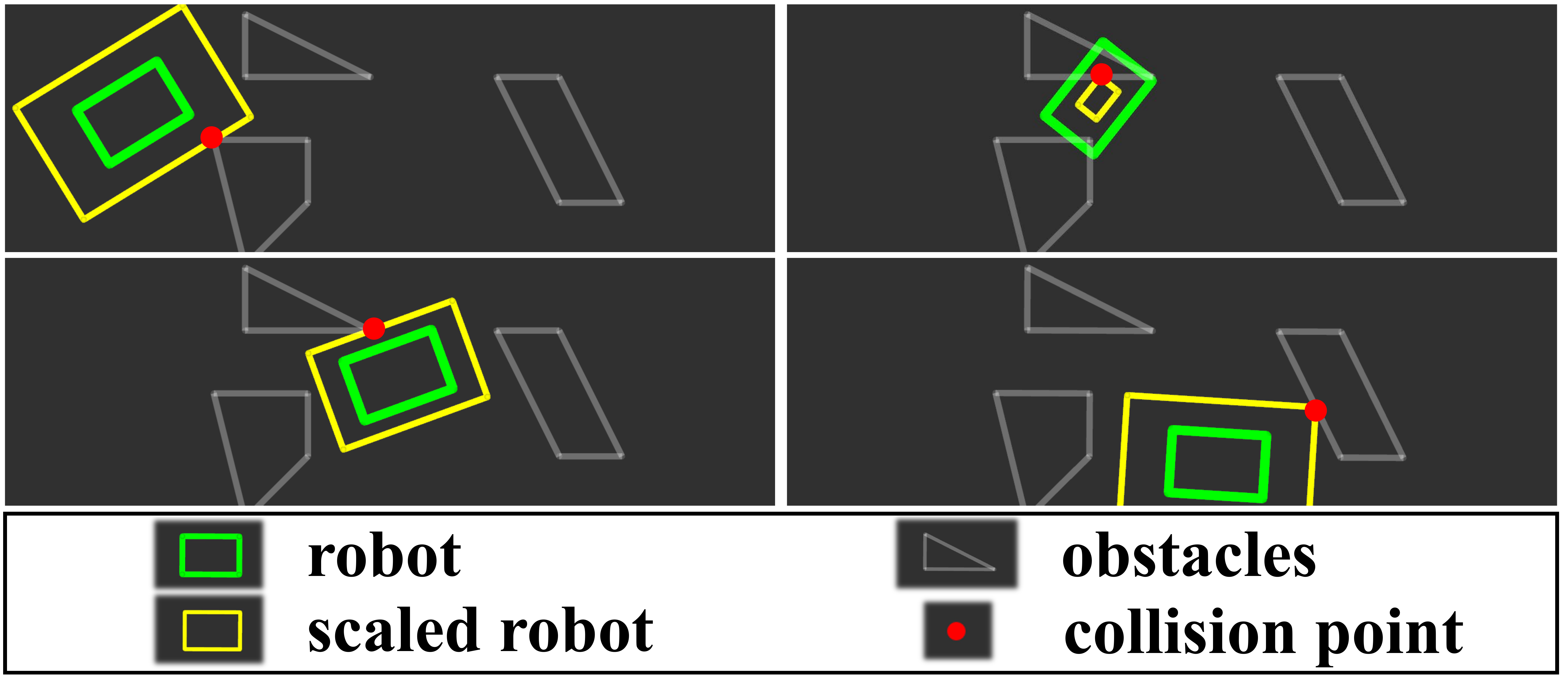}
    \vspace{-0.4cm}
	\caption{
            This figure illustrates how the robot's corresponding scaled robot changes when it moves through a map made up of some obstacles.
	}
    \vspace{-1.2cm}
	\label{fig:toutu_scale}
\end{figure}

In this paper, we propose a novel method that analytically evaluates the collisions of two convex objects, addressing the above requirements and challenges.
This method evaluates collision by calculating a minimum scale.
One of the objects enlarged or reduced by this scale can have a collision with another object, as shown in Fig.\ref{fig:toutu_scale}.
This method works seamlessly with objects represented by either points or hyperplanes, which we then abbreviate with \textbf{V}(ertex)-representation and \textbf{H}(yperplane)-representation (detailed in Sec. \ref{sec:Problem}).
Another highlight of our method is that it works directly without pre-processing even for redundant points or hyperplanes.
Then we formulate the scale calculation into a low dimensional linear programming (LP) whose computational time is $O(m)$ \cite{seidel1991small}, where $m$ is the sum of the number of points or hyperplanes that make up objects, which is also the total number of the linear inequalities in the LP.
Regardless of the number of obstacles, this method uses only one variable, the minimum scale $\beta\in \mathbb{R}_{\geq 0}$ defined in Sec.\ref{sec:Problem}, to evaluate the collision with the environment.
When applied to whole-body trajectory optimization, with the active constraints of LP, this method provides gradient of the scale $\beta$ w.r.t. the ego-motion of the scaled object analytically (detailed in Sec.~\ref{sec:Gradient}).
Even when using a naive trajectory that does not take into account the robot's shape as initial values, a safe whole-body trajectory can be obtained after optimization with our method.
Finally, to verify the generality and practicality, we apply the proposed method to generate SE(3) aerial robot trajectories based on the differential flatness of multicopter, and plan 2-d car-like robot trajectories considering nonholonomic constraints.
The major contributions of this paper are summarized as:

\begin{itemize}
    \item We propose an exact and rapid collision evaluation method that supports locality, via low-dimensional LP.
    \item We implement the method in V-representation and H-representation and derive the analytic gradient for whole-body trajectory optimization.
    \item We validate our method on several challenging cases, SE(3) aerial robots and 2-d car-like robots, testing in both static and dynamic environments.
\end{itemize}

\section{Related Work}
\label{sec:relatedworks}

\subsection{Collision Evaluation}

The simplex-based iterative approaches Gilbert-Johnson-Keerthi (GJK)  \cite{gilbert1988fast} and enhancing GJK \cite{cameron1997enhancing} have been widely utilized to calculate the distance between two convex objects.
However, these algorithms demand pre-processing to compute support functions.
When two objects intersect, the computational cost of GJK has to consider the expanding polytope algorithm (EPA)  \cite{van2003collision}.
Gilbert et al. \cite{gilbert1994new} propose a growth distance to measure collision.
Similarly, Tracy et al. \cite{tracy2022differentiable} calculate the minimum scale that both objects enlarge or reduce simultaneously for an intersection to exist.
They formulate the scale computation as a conic problem solved by primal-dual interior-point approaches, which are complex to solve and get the gradient.
Additionally, both methods do not support locality.
It means that when we demand to evaluate the collision of a robot with many obstacles, both methods require us to perform calculations with all the obstacles.
This leads to an increase in computational overhead as obstacles increases.
Moreover, the methods encounter problems with unstable values when one of the objects is much larger than the other.
Recently, Lutz et al. \cite{lutz2021efficient} propose a constructive solid geometry method based on two-layer LogSumExp functions.
But it has a gap with the true value.

\subsection{Whole-Body Trajectory Optimization}
As stated in Sec. \ref{sec:Introduction}, to achieve low-cost collision avoidance, there are several brilliant efforts in the field of robot trajectory optimization considering the whole-body shape.

Extensive works~\cite{zhou2020egoplanner,zhou2019robust} model aerial robots as spheres.
For car-like robots, Li \cite{li2021optimization} and Ziegler \cite{ziegler2014trajectory} treat with several circles intuitively.
Simply inflating the obstacles according to the robot's radius, they bound the centers of the circles or spheres in the free space of the inflate map for safety.
To generate an SE(3) trajectory that enables the drone to perch on a moving platform, Ji et al. \cite{ji2022real} model it as a disc to evaluate collision.
However, neither the conservative methods \cite{zhou2020egoplanner,zhou2019robust,li2021optimization,ziegler2014trajectory} nor the task-specific method \cite{ji2022real} can be applied in narrow environments where the physical robot shape should be considered.
Liu et al. \cite{liu2018search} propose an SE(3) planner which formulates the quadrotor as an ellipsoid to cross narrow gaps.
As the distance between obstacle and ellipsoid is hard to obtain, they adopt to generate motion primitives and check safety along every primitive. 
Nonetheless, this search-based method has to raise the primitive resolution to improve the success rate in complex environments, which leads to an explosion in computational overhead.
Zhang et al. \cite{zhang2018autonomous,zhang2020optimization} use differentiable dual variables to formulate the distance between objects to achieve optimization-based collision avoidance (OBCA).
Whereas, since the number of dual variables is related to the number of obstacles, the dimension of the problem rises considerably when the obstacles increases.
Wang \cite{wang2022geometrically} and Han \cite{han2021fast} generate safe SE(3) trajectories through optimization with explicit spatial constrains.
They generate a series of convex polyhedrons based on the free space as flight corridors and model the robot as a convex polyhedron according to its shape, ensuring trajectory safety by constraining the robot convex polyhedron in the flight corridor.
Similarly, for vehicles, Ding \cite{ding2019safe} and Manzinger \cite{manzinger2020using} construct safe corridor by rectangles.
Both the corridor-based methods require the intersection of two adjacent polyhedrons of corridor to contain at least one robot polyhedron.
However, when this demanding request is not satisfied, there is no more feasible solution.

\section{Problem Definition}
\label{sec:Problem}
We will present the problem definition of calculating the minimum scale in V-representation and H-representation respectively.
For ease of presentation, we will refer to the scaled object as the \textbf{body} and to the other object as the \textbf{obstacle} below.

\subsection{V-representation}
\label{sec:Methodology:VR}

\begin{figure}[!t]
	\centering
    \vspace{0.2cm}
	\includegraphics[width=0.95\linewidth]{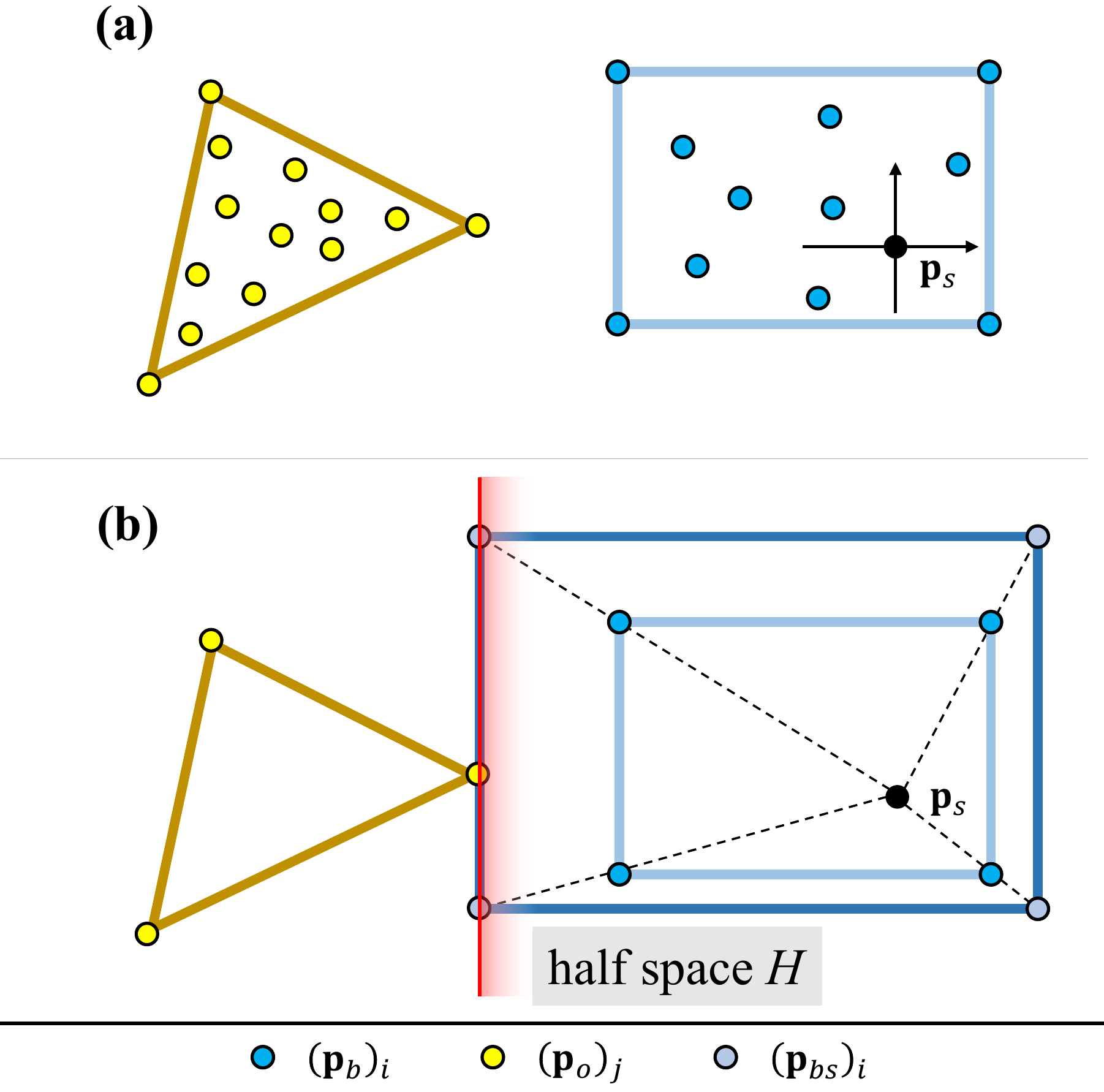}
	\caption{
            This figure illustrates the problem definition in V-representation.
            \textbf{(a)}: The obstacle and the body are defined by the yellow and blue convex hull respectively.
            We use the yellow and blue redundant points to represent the body and the obstacle respectively.
            \textbf{(b)}: To prevent confusion in the visualisation, the redundant points are not illustrated.
            The light blue points represent the scaled body and the red line denotes the half space containing the scaled body but not the obstacle.
	}
	\label{fig:vr_demo}
    \vspace{-0.8cm}
\end{figure}

In V-representation, as shown in Fig.\ref{fig:vr_demo}{(a)}, both the yellow and blue objects are defined by a convex hull that can contain the redundant points.
However, instead of processing points into a convex hull, the proposed method is capable of working directly on the redundant points. 
For the simplicity of visualization, we do not visualize points that are redundant for representing object in Fig.\ref{fig:vr_demo}{(b)}.

As illustrated in Fig.\ref{fig:vr_demo}{(a)}, we use point sets $P_{body}^{b}=\{(\mathbf{p}_b^{b})_i|i=1,2,...,n_b\}$ and $P_{obs}^{b}=\{(\mathbf{p}_o^{b})_j|j=1,2,...,n_o\}$ to represent body and obstacle in the body frame.
We define a point $\mathbf{p}_s$ in the body frame as scale seed point, which the body point sets scale about.
We define $\beta\in \mathbb{R}_{\geq 0}$ as the scale.
Then we get obstacle and scaled body point sets in the coordinate system with the point $\mathbf{p}_s$ as the origin as 
\begin{equation}
\label{eq:scale_eq}
\begin{aligned}
    P_{body}^{s}(\beta)&=\{(\mathbf{p}_{bs}^{s})_i=\beta \left( (\mathbf{p}_b^{b})_i-\mathbf{p}_s^{b} \right)|i=1,2,...,n_b\},\\
    P_{obs}^{s}&=\{(\mathbf{p}_{o}^{s})_j=(\mathbf{p}_o^{b})_j-\mathbf{p}_s^{b}|j=1,2,...,n_o\},
\end{aligned}
\end{equation}

Then we define the problem of calculating the minimum scale in V-representation to maximize the scale $\beta$ with the constraints of
\begin{equation}
\label{eq:constrains}
    \left\{
    \begin{aligned}
        \bm{\alpha}_o\tp\beta \left( (\mathbf{p}_b^{b})_i-\mathbf{p}_s^{b} \right)&\leq 1, &i=1,2,...,n_b \\
        \bm{\alpha}_o\tp \left( (\mathbf{p}_o^{b})_j-\mathbf{p}_s^{b} \right) &\geq 1, &j=1,2,...,n_o
    \end{aligned}
    \right.,
\end{equation}
which means a half space $H=\{x|\bm{\alpha}_o\tp x\leq1\}$ is required so that the scaled body $P_{body}^{s}(\beta)$ is inside the half space and the obstacle $P_{obs}^{s}$ is outside, as shown in Fig.\ref{fig:vr_demo}{(b)}.

We define $\bm{\alpha}\tp = \beta\bm{\alpha}_o\tp$.
Since $\beta\in \mathbb{R}_{\geq 0}$, we can formulate the scale calcluation as a low dimension LP problem:
\begin{equation}
\label{eq:v_sdlp}
\begin{split}
&\max \beta\\
&s.t.\quad  \left\{
    \begin{array}{lc}
        \bm{\alpha}\tp \left( (\mathbf{p}_b^{b})_i-\mathbf{p}_s^{b} \right)\leq 1, ~&i=1,2,...,n_b \\
        \bm{\alpha}\tp \left( (\mathbf{p}_o^{b})_j-\mathbf{p}_s^{b} \right) \geq \beta , ~&j=1,2,...,n_o
    \end{array}\right..
\end{split}
\end{equation}

\begin{figure}[!t]
	\centering
    \vspace{0.2cm}
	\includegraphics[width=0.95\linewidth]{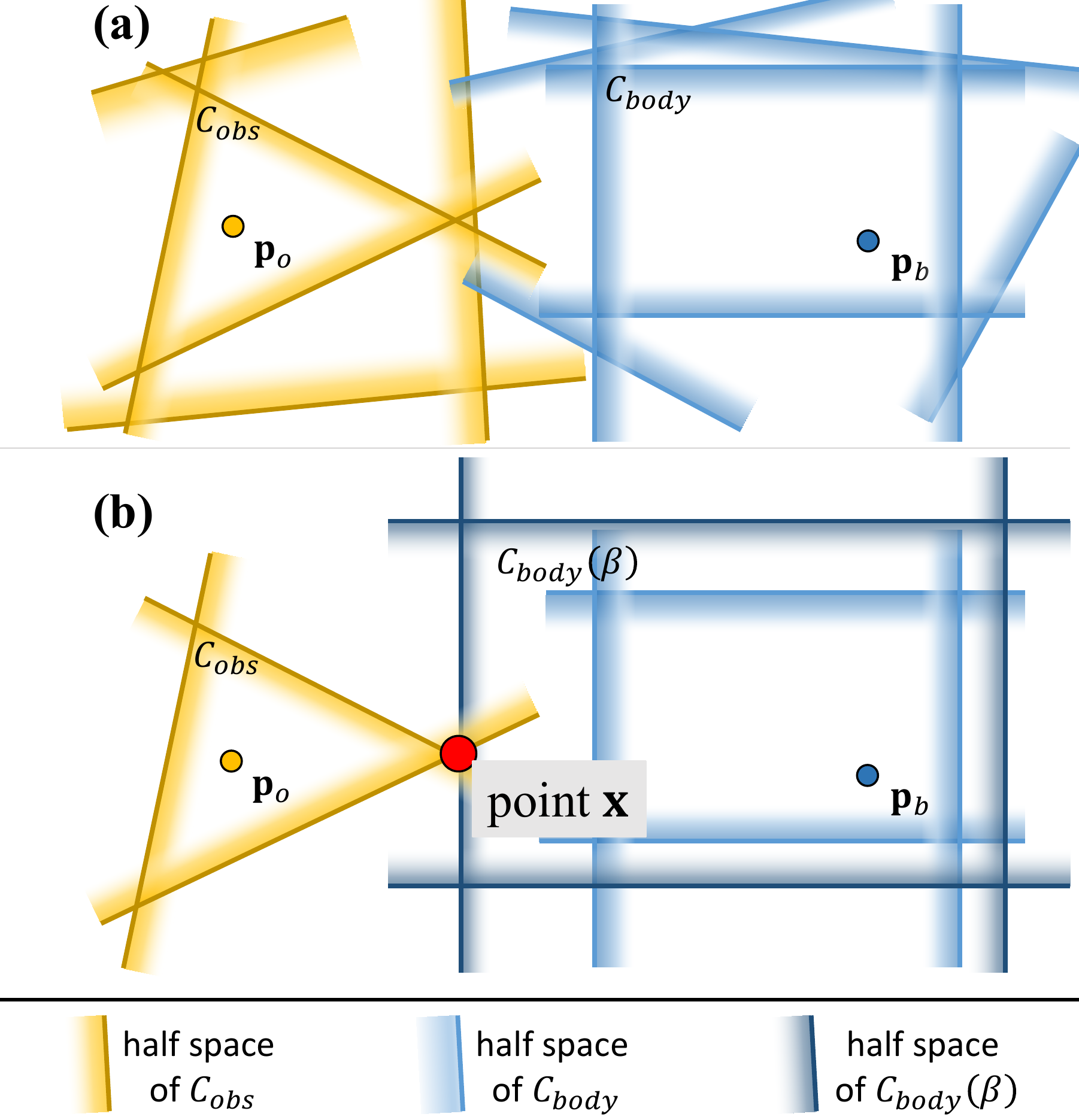}
	\caption{
        This figure illustrates the problem definition in H-representation.
        \textbf{(a)}: The obstacle and the body are represented by intersections of several redundant yellow and blue half spaces respectively.
        \textbf{(b)}: For better visualization, we do not show the half spaces that are redundant for the representation.
        The dark blue points represent the scaled body and the red point denotes the point that belongs to both the scaled body and the obstacle.
	}
	\label{fig:hr_demo}
    \vspace{-0.5cm}
\end{figure}

\subsection{H-representation}
\label{sec:Methodology:HR}
In H-representation,  as illustrated in Fig.\ref{fig:hr_demo}{(a)}, we use the intersection of several redundant half spaces to represent a convex polyhedron. 
For a clear visualization, the redundant half spaces are not shown in Fig.\ref{fig:hr_demo}{(b)}.

As indicated in Fig.\ref{fig:hr_demo}{(a)}, we use $C_{body}=\{\mathbf{x}|(\bm{\alpha}_{b})_i\tp (\mathbf{x}-\mathbf{p}_b) \leq 1, i=1,2,...,n_b\}$ and $C_{obs}=\{\mathbf{x}|(\bm{\alpha}_{o})_j\tp (\mathbf{x}-\mathbf{p}_o) \leq 1, j=1,2,...,n_o\}$ to represent body and obstacle in the world frame.
$\mathbf{p}_b$ and $\mathbf{p}_o$ are points inside body and obstacle respectively.
$n_b$ and $n_o$ are the numbers of half spaces whose intersections define body and obstacle.
We make the body scale about $\mathbf{p}_b$, then the scaled body can be written as
\begin{equation}
    C_{body}(\beta)=\{\mathbf{x}|(\bm{\alpha}_{b})_i\tp (\mathbf{x}-\mathbf{p}_b) \leq \beta, i=1,2,...,n_b\},
\end{equation}
where $\beta\in \mathbb{R}_{\geq 0}$ is the scale.

Referring to Eq.\ref{eq:constrains} and Eq.\ref{eq:v_sdlp}, we define the problem in H-representation as 
\begin{equation}
\label{eq:h_sdlp}
\begin{split}
&\min \beta\\
&s.t.\quad  \left\{
    \begin{array}{lc}
        (\bm{\alpha}_{b})_i\tp (\mathbf{x}-\mathbf{p}_b)\leq \beta, ~&i=1,2,...,n_b \\
        (\bm{\alpha}_{o})_j\tp (\mathbf{x}-\mathbf{p}_o) \leq 1 , ~&j=1,2,...,n_o
    \end{array}\right.,
\end{split}
\end{equation}
which is obviously a low dimension LP problem.
And the constraints of Eq.\ref{eq:h_sdlp} means that a point $\mathbf{x}$ which satisfies $\mathbf{x} \in C_{body}(\beta) \cap C_{obs}$ is required as illustrated in Fig.\ref{fig:hr_demo}{(b)}.

\section{Gradient Computation}
\label{sec:Gradient}

Since both problems we defined in V-representation and H-representation are in the form of LP, we only derive the sub-gradient in V-representation in this paper and the idea of gradient computation in H-representation is similar.

We use the active constraints of the low dimension LP problem to get the gradient of the scale $\beta$. 
In $n$-dimensional environment, based on the LP problem definition in Eq.\ref{eq:v_sdlp}, there are two different kinds of active constraints which can be written in the form of linear equations:
\begin{equation}
\label{eq:ac_le}
\begin{aligned}
    \begin{bmatrix}
        \left( (\mathbf{p}_{b\_ac}^{b})_i-\mathbf{p}_s^{b} \right)\tp ~~~~~0 
    \end{bmatrix}
    \begin{bmatrix}
        \bm{\alpha}_{n \times 1} \\ \beta
    \end{bmatrix}&=1,  ~i=1,2,...,n_{b\_ac} \\
    \begin{bmatrix}
        \left( (\mathbf{p}_{o\_ac}^{b})_j-\mathbf{p}_s^{b} \right)\tp ~-1 
    \end{bmatrix}
    \begin{bmatrix}
        \bm{\alpha}_{n \times 1} \\ \beta
    \end{bmatrix}&=0,  ~j=1,2,...,n_{o\_ac}
\end{aligned}
\end{equation}
where $(\mathbf{p}_{b\_ac}^{b})_i \in P_{body}^{b}$, $(\mathbf{p}_{o\_ac}^{b})_j \in P_{obs}^{b}$ and
$n_{b\_ac}+n_{o\_ac}=n+1$, which means we should have $n+1$ active constraints to solve the ($n+1$)-dimensional LP problem. 

\begin{figure}[H]
	\centering
    \vspace{0.2cm}
	\includegraphics[width=0.9\linewidth]{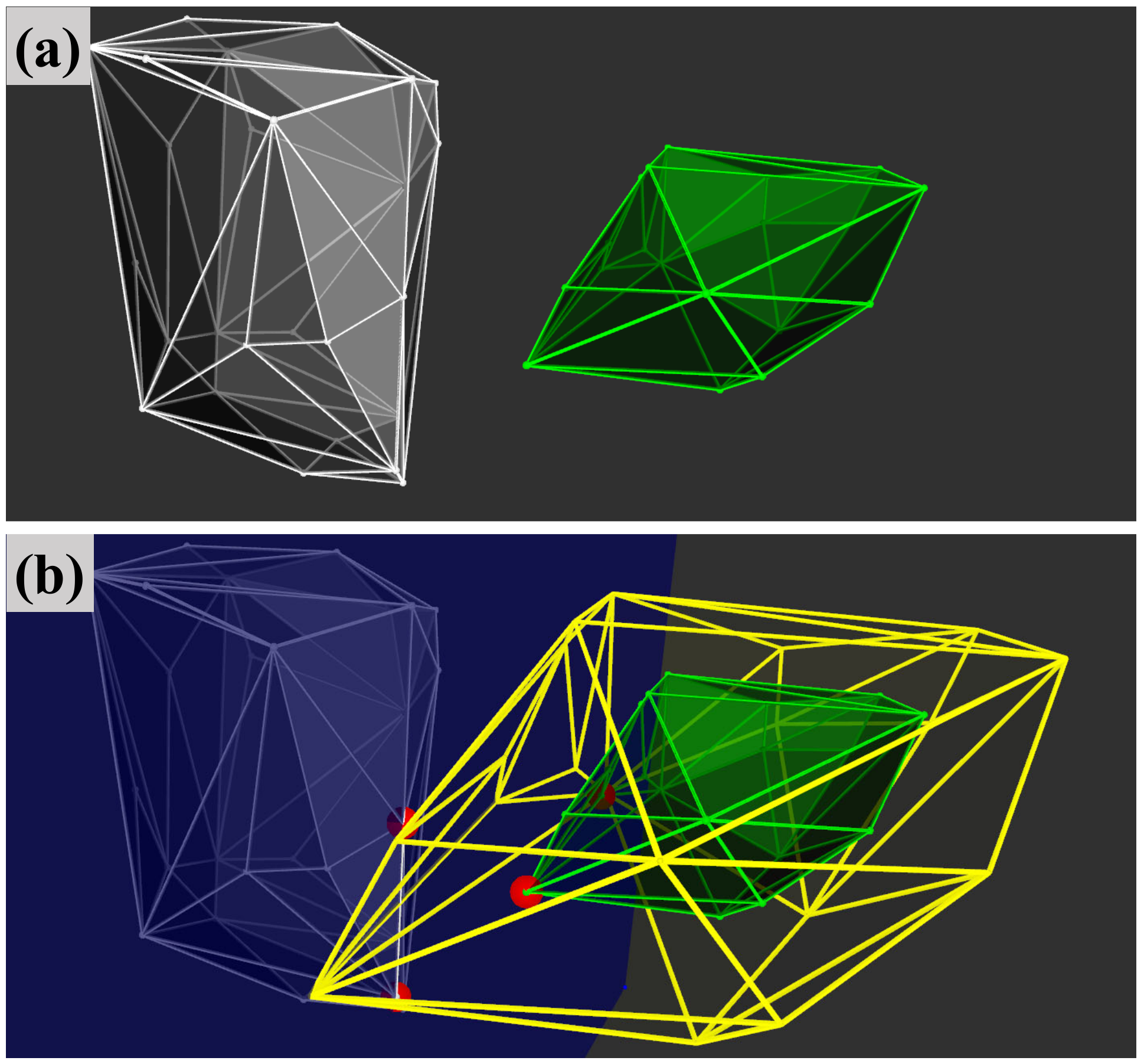}
	\vspace{-0.2cm}
	\caption{
        Illustration of calculation of the minimum scale in V-representation.
        \textbf{(a)}: The white and the green convex hull denote the obstacle and the body respectively.
        \textbf{(b)}: The yellow convex hull is the scaled body obtained by enlarging the body to the minimum scale.
        The blue plane is the hyperplane splitting the scaled body and the obstacle, and the red points represent the active constraint points that satisfy one of the linear equations in Eq.\ref{eq:ac_le}.
        Although the objects are visualized using convex hull, the method can be applied directly to the redundant point sets during the calculation.
	}
	\label{fig:gradient}
	\vspace{-0.2cm}
\end{figure}

We refer to $(\mathbf{p}_{b\_ac}^{b})_i$ and $(\mathbf{p}_{o\_ac}^{b})_j$ as the active constraint point on the body and the obstacle respectively.
In Fig.\ref{fig:gradient}, We show the result of the minimum scale calculation defined in V-representation.
As indicated in Fig.\ref{fig:gradient}{(b)}, the red points in the green and white convex hull are the $(\mathbf{p}_{b\_ac}^{b})_i$ and $(\mathbf{p}_{o\_ac}^{b})_j$ respectively.
Additionally, the points in the scaled body obtained from the $(\mathbf{p}_{b\_ac}^{b})_i$ according to Eq.\ref{eq:scale_eq} and the points $(\mathbf{p}_{o\_ac}^{b})_j$ are on the hyperplane $P=\{x|\bm{\alpha}\tp x = \beta\}$ which is represented in Fig.\ref{fig:gradient}{(b)} by the blue plane.

We combine all the linear equations in Eq.\ref{eq:ac_le} and write them in matrix form as
\begin{equation}
\label{eq:ac_matrix}
    \begin{bmatrix}
        \begin{matrix}
            \left( (\mathbf{p}_{b\_ac}^{b})_1-\mathbf{p}_s^{b} \right)\tp \\ 
            \vdots \\ 
            \left( (\mathbf{p}_{b\_ac}^{b})_{n_{b\_ac}}-\mathbf{p}_s^{b} \right)\tp \\
            \left( (\mathbf{p}_{o\_ac}^{b})_1-\mathbf{p}_s^{b} \right)\tp \\ 
            \vdots \\ 
            \left( (\mathbf{p}_{o\_ac}^{b})_{n_{o\_ac}}-\mathbf{p}_s^{b} \right)\tp 
        \end{matrix}
        &
        \begin{matrix}
            0 \\ \vdots \\ 0 \\
            -1 \\ \vdots \\ -1
        \end{matrix}\\
    \end{bmatrix}
    \begin{bmatrix}
        \bm{\alpha}_{n \times 1} \\ \beta
    \end{bmatrix}=
    \begin{bmatrix}
        1 \\ \vdots \\ 1 \\ 
        0 \\ \vdots \\ 0
    \end{bmatrix}.
\end{equation}
Then we block this matrix equation in Eq.\ref{eq:ac_matrix} in into
\begin{equation}
\label{eq:matrix_solution}
    \begin{bmatrix}
        \mathbf{A}_{n\times n} ~~\mathbf{B}_{n\times 1}\\
        \mathbf{C}_{1\times n} ~~\mathbf{D}_{1\times 1}\\
    \end{bmatrix}
    \begin{bmatrix}
        \bm{\alpha}_{n \times 1} \\ \beta
    \end{bmatrix}=
    \begin{bmatrix}
        \mathbf{E}_{n\times 1} \\ \mathbf{F}_{1\times 1}
    \end{bmatrix},
\end{equation}
which can be written into a system of linear equations with two variables,
where the variable $\bm{\alpha}_{n \times 1}$ can be eliminated and get an equation which only has one variable $\beta$:
\begin{equation}
\label{eq:solved_matrix}
    \mathbf{C}_{1\times n} \mathbf{A}_{n\times n}^{-1}(\mathbf{E}_{n\times 1}-\mathbf{B}_{n\times 1}\beta)+\mathbf{D}_{1\times 1}\beta = \mathbf{F}_{1\times 1}.
\end{equation}

For a rigid body, the point set $P_{body}^{b}$ in body frame is not related to the motion of the body. 
As for $P_{obs}^{b}$, in practice we can only directly obtain the obstacle's point set $P_{obs}^{w}$ in world frame.
In order not to lose generality, we define the center of rotation of the rigid body in the body frame as $\mathbf{p}_{cen}^{b}$, and define $\mathbf{R}$ and $\mathbf{t}$ for the rotation and translation of the body in the world frame.
Based on the body's motion and $P_{obs}^{w}$, we can get the point in $P_{obs}^{b}$ as
\begin{equation}
\label{eq:motion}
    (\mathbf{p}_{o}^{b})_j=\mathbf{R}^{-1} \left( (\mathbf{p}_o^{w})_j-\mathbf{t}-\mathbf{p}_{cen}^{b} \right),
\end{equation}
 
Then we can conveniently use the implicit function in Eq.\ref{eq:solved_matrix}, which contains the relationship between the scale and the motion of the rigid body, to obtain the partial derivatives of $\beta$ w.r.t. $\mathbf{R}$ and $\mathbf{t}$.
When $\mathbf{R}$ and $\mathbf{t}$ are time-dependent, such as using a time-parameterized polynomial trajectory to represent motion, the gradient of $\beta$ w.r.t. time can also be obtained.

\section{Application on Aerial Robots}
\label{sec:application}

To demonstrate the versatility of our method on aerial robots, we apply it to a whole-body multicopter trajectory optimization problem, based on the differential flatness of multicopter in Wang's work \cite{wang2022geometrically}.
As shown in Fig.\ref{fig:toutu}, we model the multicopter with a green rectangle that measures $3 \times 2 \times 0.6~m$, and require it to fly through a narrow slit.
We adopt a segmented polynomial to represent the flat-output trajectory, of which we use the MINCO \cite{wang2022geometrically} to conduct spatial-temporal deformation.
In this SE(3) trajectory optimization, we set the maximum velocity $6m/s$ and acceleration $10m/s^2$, considering smoothness, safety and dynamic feasibility simultaneously.
Moreover, we achieve obstacle avoidance by constraining the minimum scale defined in Sec.~\ref{sec:Problem} greater than 1, which needs the gradient of the scale w.r.t. the motion of the multicopter.

\subsection{Gradient Computation for SE(3) Motion}

We derive the calculation of the gradient in detail.
As defined in Sec. \ref{sec:Gradient}, we use $\mathbf{R}$ and $\mathbf{t}$ to represent the rotation matrix and translation of the body.
For convenience, in optimization we use a normalized quaternion $\mathbf{q} = [w, x, y, z]\tp$ to represent rotation.
Referring to \cite{sola2017quaternion}, the rotation matrix $\mathbf{R}$ can be expressed  by the quaternion $\mathbf{q}$ and the partial derivatives $\frac{\partial \mathbf{R}}{\partial \mathbf{q}.*},*=\{w, x, y, z\}$ can be easily obtained.

In this case where $n=3$, as mentioned in Sec. \ref{sec:Gradient}, the LP problem which defined in Sec. \ref{sec:Methodology:VR} should have $4$ active constraints. 
The specific situations of active constraints (\textbf{ac}) can be divided into three types:
\begin{itemize}
    \item 3 \textbf{ac} $\in P_{body}^{b}$, 1 \textbf{ac} $\in P_{obs}^{b}$;
    \item 2 \textbf{ac} $\in P_{body}^{b}$, 2 \textbf{ac} $\in P_{obs}^{b}$;
    \item 1 \textbf{ac} $\in P_{body}^{b}$, 3 \textbf{ac} $\in P_{obs}^{b}$.
\end{itemize}

We then analyse each type in turn:

\subsubsection{ 3 \textbf{ac} $\in P_{body}^{b}$, 1 \textbf{ac} $\in P_{obs}^{b}$}
In this case, the block matrix equation in Eq.\ref{eq:matrix_solution} can be written as
\begin{equation}
    \begin{bmatrix}
        \begin{bmatrix}
            \left( (\mathbf{p}_{b\_ac}^{b})_1-\mathbf{p}_s^{b} \right)\tp \\ 
            \left( (\mathbf{p}_{b\_ac}^{b})_2-\mathbf{p}_s^{b} \right)\tp \\ 
            \left( (\mathbf{p}_{b\_ac}^{b})_3-\mathbf{p}_s^{b} \right)\tp 
        \end{bmatrix}
        &
        \begin{matrix}
            0 \\ 0 \\ 0 
        \end{matrix}\\
        \left( (\mathbf{p}_{o\_ac}^{b})_1-\mathbf{p}_s^{b} \right)\tp & -1
    \end{bmatrix}
    \begin{bmatrix}
        \bm{\alpha}_{3\times1} \\ \beta
    \end{bmatrix}=
    \begin{bmatrix}
        1 \\ 1 \\ 1 \\ 0
    \end{bmatrix}.
\end{equation}
Then based on Eq.\ref{eq:solved_matrix}, we can get
\begin{equation}
\label{eq:31beta}
    \beta=\left( (\mathbf{p}_{o\_ac}^{b})_1-\mathbf{p}_s^{b} \right)\tp
    \begin{bmatrix}
        \left( (\mathbf{p}_{b\_ac}^{b})_1-\mathbf{p}_s^{b} \right)\tp \\ 
        \left( (\mathbf{p}_{b\_ac}^{b})_2-\mathbf{p}_s^{b} \right)\tp \\ 
        \left( (\mathbf{p}_{b\_ac}^{b})_3-\mathbf{p}_s^{b} \right)\tp 
    \end{bmatrix}^{-1}
    \begin{bmatrix}
        1 \\ 1 \\ 1 
    \end{bmatrix},
\end{equation}
where only $(\mathbf{p}_{o\_ac}^{b})_1$ is related to $\mathbf{R}$ and $\mathbf{t}$. 
Based on Eq.\ref{eq:motion}, for brevity, we write this equation in Eq.\ref{eq:31beta} as
\begin{equation}
    \label{eq:31beta_simple}
        \beta
        =\left( (\mathbf{p}_{o\_ac}^{b})_1-\mathbf{p}_s^{b} \right)\tp
        \mathbf{A}^{-1} \mathbf{E},
    \end{equation}
where $\mathbf{A}$ and $\mathbf{E}$, defined in Eq.\ref{eq:matrix_solution}, are constant in this case.
Then we can get the gradient of $\beta~~w.r.t.~~\mathbf{t}$ and $\mathbf{q}$ as
\begin{equation}
    \frac{\partial \beta}{\partial \mathbf{t}}=
    -\mathbf{R}
    \mathbf{A}^{-1} \mathbf{E},~~
    \frac{\partial \beta}{\partial \mathbf{q}.*}=-
    (\mathbf{p}_{o\_ac}^{b})_1 \tp
    \frac{\partial \mathbf{R}\tp}{\partial \mathbf{q}.*}
    \mathbf{R}
    \mathbf{A}^{-1} \mathbf{E}.
\end{equation}


\subsubsection{2 \textbf{ac} $\in P_{body}^{b}$, 2 \textbf{ac} $\in P_{obs}^{b}$}
In this case, the block matrix equation in Eq.\ref{eq:matrix_solution} can be written as
\begin{equation}
    \begin{bmatrix}
        \begin{bmatrix}
            \left( (\mathbf{p}_{b\_ac}^{b})_1-\mathbf{p}_s^{b} \right)\tp \\ 
            \left( (\mathbf{p}_{b\_ac}^{b})_2-\mathbf{p}_s^{b} \right)\tp \\ 
            (\Delta \mathbf{p}_{o\_ac}^{b})\tp 
        \end{bmatrix}
        &
        \begin{matrix}
            0 \\ 0 \\ 0 
        \end{matrix}\\
        \left( (\mathbf{p}_{o\_ac}^{b})_1-\mathbf{p}_s^{b} \right)\tp & -1
    \end{bmatrix}
    \begin{bmatrix}
        \bm{\alpha}_{3\times1} \\ \beta
    \end{bmatrix}=
    \begin{bmatrix}
        1 \\ 1 \\ 0 \\ 0
    \end{bmatrix},
\end{equation}
where the \textbf{ac} corresponding to $\Delta \mathbf{p}_{o\_ac}^{b}$ is
\begin{equation}
\begin{aligned}
    \left( (\mathbf{p}_{o\_ac}^{b})_1-\mathbf{p}_s^{b} \right)\tp \bm{\alpha} - \beta = 0,\\
    \left( (\mathbf{p}_{o\_ac}^{b})_2-\mathbf{p}_s^{b} \right)\tp \bm{\alpha} - \beta = 0,\\
\end{aligned}
\end{equation}
which can be combined to obtain
\begin{equation}
    \left((\mathbf{p}_{o\_ac}^{b})_1\tp - (\mathbf{p}_{o\_ac}^{b})_2\tp \right) \bm{\alpha}
    = (\Delta\mathbf{p}_{o\_ac}^{b})\tp \bm{\alpha}
    = 0.
\end{equation}
Based on Eq.\ref{eq:motion}, $\Delta \mathbf{p}_{o\_ac}^{b}$ can be written as 
\begin{equation}
    \Delta\mathbf{p}_{o\_ac}^{b}=\mathbf{R}^{-1} \left( (\mathbf{p}_o^{w})_1-(\mathbf{p}_o^{w})_2 \right).
\end{equation}
Then based on Eq.\ref{eq:solved_matrix}, we can get
\begin{equation}
    \beta=
    \left( (\mathbf{p}_{o\_ac}^{b})_1-\mathbf{p}_s^{b} \right)\tp
    \begin{bmatrix}
        \left( (\mathbf{p}_{b\_ac}^{b})_1-\mathbf{p}_s^{b} \right)\tp \\ 
        \left( (\mathbf{p}_{b\_ac}^{b})_2-\mathbf{p}_s^{b} \right)\tp \\ 
        (\Delta \mathbf{p}_{o\_ac}^{b})\tp 
    \end{bmatrix}^{-1}
    \begin{bmatrix}
        1 \\ 1 \\ 0
    \end{bmatrix},
\end{equation}
which, for brevity, we write as
\begin{equation}
    \beta=
    \mathbf{C} \mathbf{A}^{-1} \mathbf{E},
\end{equation}
where $\mathbf{C}$, $\mathbf{A}$ and $\mathbf{E}$ are defined in Eq.\ref{eq:matrix_solution}.
In this case, $\mathbf{C}$ is related to $\mathbf{R}$ and $\mathbf{t}$, $\mathbf{A}$ is only related to $\mathbf{R}$, and $\mathbf{E}$ is constant.

Then we can get the gradient of $\beta~~w.r.t.~~\mathbf{t}$ as
\begin{equation}
    \frac{\partial \beta}{\partial \mathbf{t}}=
    -\mathbf{R}  \mathbf{A}^{-1} \mathbf{E}.
\end{equation}

And we can get the gradient of $\beta~~w.r.t.~~\mathbf{q}$ as
\begin{equation}
\begin{aligned}
    \frac{\partial \beta}{\partial \mathbf{q}.*}=&-
    (\mathbf{p}_{o\_ac}^{b})_1 \tp
    \frac{\partial \mathbf{R} \tp}{\partial \mathbf{q}.*}
    \mathbf{R}\mathbf{A}^{-1} \mathbf{E}\\
    &+\mathbf{C}\mathbf{A}^{-1} 
    \begin{bmatrix}
        \mathbf{0}_{1\times3} \\ \mathbf{0}_{1\times3} \\ 
        (\Delta\mathbf{p}_{o\_ac}^{b})\tp
        \frac{\partial \mathbf{R} \tp}{\partial \mathbf{q}.*}\mathbf{R}
    \end{bmatrix}
    \mathbf{A}^{-1} \mathbf{E}.
\end{aligned}
\end{equation}

\subsubsection{ 1 \textbf{ac} $\in P_{body}^{b}$, 3 \textbf{ac} $\in P_{obs}^{b}$}
In this case, the block matrix equation in Eq.\ref{eq:matrix_solution} can be written as
\begin{equation}
    \begin{bmatrix}
        \begin{bmatrix}
            \left( (\mathbf{p}_{o\_ac}^{b})_1-\mathbf{p}_s^{b} \right)\tp \\
            \left( (\mathbf{p}_{o\_ac}^{b})_2-\mathbf{p}_s^{b} \right)\tp \\
            \left( (\mathbf{p}_{o\_ac}^{b})_3-\mathbf{p}_s^{b} \right)\tp
        \end{bmatrix}
        &
        \begin{matrix}
            -1 \\ -1 \\ -1 
        \end{matrix}\\
        \left( (\mathbf{p}_{b\_ac}^{b})_1-\mathbf{p}_s^{b} \right)\tp & 0
    \end{bmatrix}
    \begin{bmatrix}
        \bm{\alpha}_{3\times1} \\ \beta
    \end{bmatrix}=
    \begin{bmatrix}
        0 \\ 0 \\ 0 \\ 1 
    \end{bmatrix}.
\end{equation}

Then based on Eq.\ref{eq:solved_matrix}, we can get
\begin{equation}
\label{eq:1b3o_problem}
    -\left( (\mathbf{p}_{b\_ac}^{b})_1-\mathbf{p}_s^{b} \right)\tp
    \begin{bmatrix}
        \left( (\mathbf{p}_{o\_ac}^{b})_1-\mathbf{p}_s^{b} \right)\tp \\
        \left( (\mathbf{p}_{o\_ac}^{b})_2-\mathbf{p}_s^{b} \right)\tp \\
        \left( (\mathbf{p}_{o\_ac}^{b})_3-\mathbf{p}_s^{b} \right)\tp
    \end{bmatrix}^{-1}
    \begin{bmatrix}
        -1 \\ -1 \\ -1 
    \end{bmatrix}\beta=1,
\end{equation}
which, for brevity, we write as
\begin{equation}
\label{eq:1b3o_problem_simple}
    -\mathbf{C} \mathbf{A}^{-1}
    \mathbf{B}
    \beta=1,
\end{equation}
where $\mathbf{C}$, $\mathbf{A}$ and $\mathbf{B}$ are defined in Eq.\ref{eq:matrix_solution}.
$\mathbf{C}$ and $\mathbf{B}$ are constant in this case.
Only $\mathbf{A}$ is related to $\mathbf{R}$ and $\mathbf{t}$.
 
Then we can get the gradient of $\beta$ with respect to $\mathbf{t}=\{\mathbf{t}.x, \mathbf{t}.y, \mathbf{t}.z\}$, as
\begin{equation}
    \mathbf{C} \mathbf{A}^{-1}
    \mathbf{B}
    \frac{\partial \beta}{\partial \mathbf{t}.x}+
    \mathbf{C} \mathbf{A}^{-1} 
    \begin{bmatrix}
        1&0&0\\
        1&0&0\\
        1&0&0\\
    \end{bmatrix}
    \mathbf{R}
    \mathbf{A}^{-1}\mathbf{B}
    \beta = 0.
\end{equation}

Based Eq.\ref{eq:1b3o_problem_simple}, we can obtain the gradient as
\begin{equation}
    \frac{\partial \beta}{\partial \mathbf{t}}=\beta
    \mathbf{R}
    \mathbf{A}^{-1}\mathbf{B}.
\end{equation}

And we can get the gradient of $\beta~~w.r.t.~~\mathbf{q}$ as
\begin{equation}
    \frac{1}{\beta}
    \frac{\partial \beta}{\partial \mathbf{q}.*}-
    \mathbf{C} \mathbf{A}^{-1} 
    \frac{\partial \mathbf{A}}{\partial \mathbf{q}.*}
    \mathbf{A}^{-1}\mathbf{B}
    \beta = 0,
\end{equation}
which can be written as
\begin{equation}
    \frac{\partial \beta}{\partial \mathbf{q}.*}=
    \mathbf{C} \mathbf{A}^{-1} 
    \frac{\partial \mathbf{A}}{\partial \mathbf{q}.*}
    \mathbf{A}^{-1}\mathbf{B}
    \beta^2,
\end{equation}
where $\frac{\partial \mathbf{A}}{\partial \mathbf{q}.*}$ can be get by
\begin{equation}
    \frac{\partial \mathbf{A}}{\partial \mathbf{q}.*}=-
    \begin{bmatrix}
         (\mathbf{p}_{o\_ac}^{b})_1 \tp \\
         (\mathbf{p}_{o\_ac}^{b})_2 \tp \\
         (\mathbf{p}_{o\_ac}^{b})_3 \tp
    \end{bmatrix}
    \frac{\partial \mathbf{R} \tp}{\partial \mathbf{q}.*}
    \mathbf{R},
\end{equation}

\subsection{Experiment Result}
L-BFGS\footnote{https://github.com/ZJU-FAST-Lab/LBFGS-Lite} \cite{liu1989limited} is adopted as an efficient
quasi-Newton method to solve the numerical optimization problem. 
We use Lewis-Overton line search \cite{lewis2013nonsmooth} to deal with the nonsmoothness of the scale, which sometimes occurs during optimization.
As the optimization result shows in Fig.\ref{fig:toutu}, the SE(3) whole-body trajectory generated by our method is collision-free and smooth.


\begin{figure}[ht]
	\centering
    \vspace{0.2cm}
	\includegraphics[width=1\linewidth]{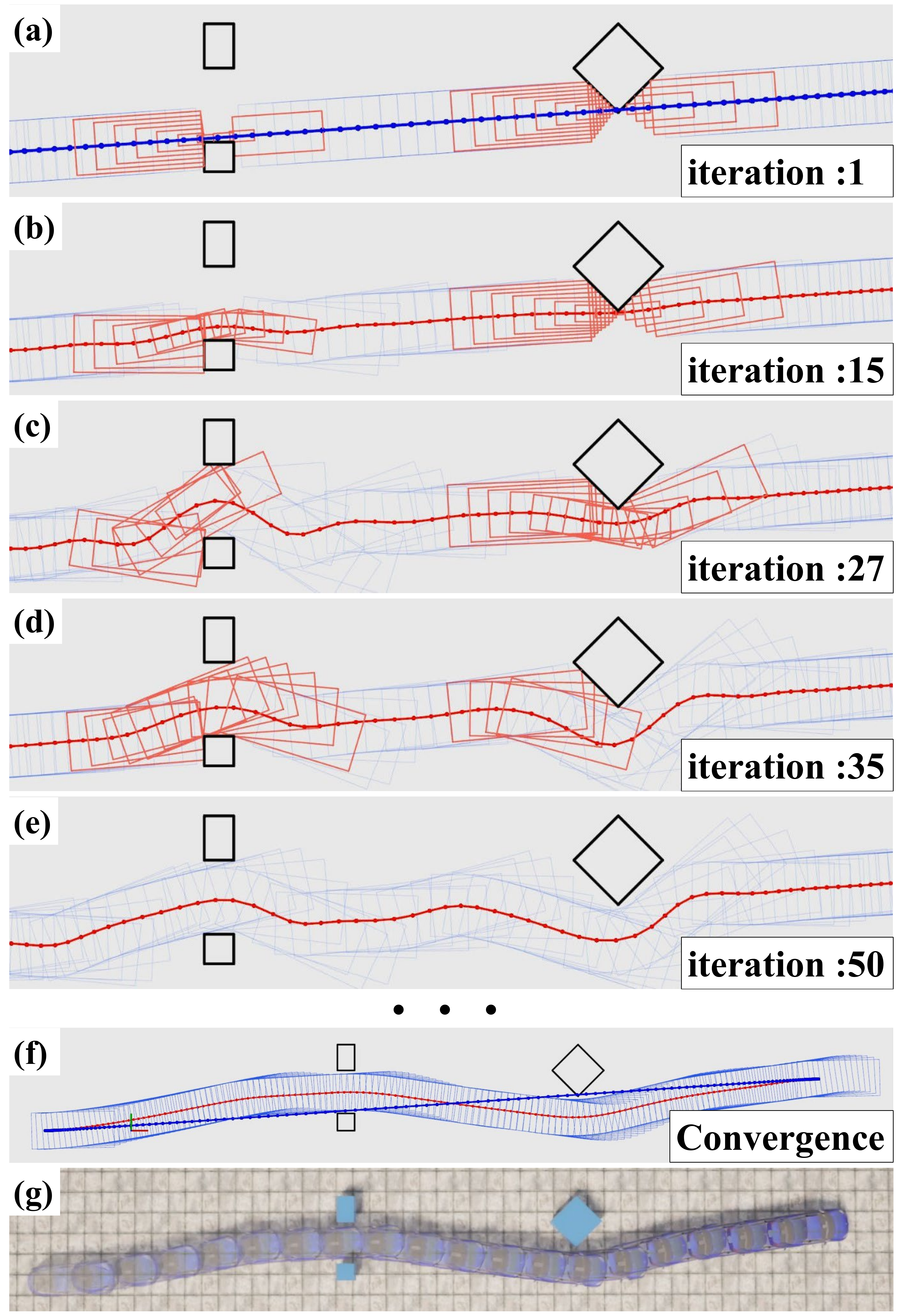}
    \vspace{-0.7cm}
	\caption{
        Illustration of the trajectory optimization.
        The blue curve indicates the initial trajectory, the red curve is the trajectories in optimization.
        \textbf{(a)-(e)}: Red squares indicate states on trajectories whose scale is small than 1, the light blue squares indicate the states that meet the scale constraint.
        \textbf{(f)}: The converged trajectory is shown, as there are no states on the trajectory that violate the scale constraint after \textbf{(e)}.
        \textbf{(g)}: The snapshot of result in CARLA.
	}
	\label{fig:car_optimization}
    \vspace{-0.6cm}
\end{figure}


\section{Application on Car-like Robots}
\label{sec:application_car}
To demonstrate the applicability of our approach to car-like robots.
As shown in Fig.\ref{fig:car_optimization}{(g)} and Fig.\ref{fig:car}, we perform the experiments in the physical simulator CARLA~\cite{dosovitskiy2017carla}.
Similar to the application in Sec. \ref{sec:application}, we adopt a segmented polynomial deformed by MINCO~\cite{wang2022geometrically} to represent the blue vehicle's trajectory and L-BFGS to solve the numerical problem of the optimization.
The trajectory optimization simultaneously considers smoothness, safety and dynamic feasibility.
The difference, however, is that the trajectory is 2-d and the nonholonomic constraints are taken into account in this application.
Additionally, due to the dynamic environment in Fig.\ref{fig:car}, we constrain the minimum scale greater than~1 in the spatial-temporal trajectory optimization for safety.
(In practice, we set the minimum scale to 1.1 for greater security.)
The gradient of the scale w.r.t. the 2-d motion is similar to that in SE(3) and will not be detailed here.

\subsection{Experiment in Static Environment}

As shown in Fig.\ref{fig:car_optimization}{(g)}, we apply our method to the blue vehicle's whole-body trajectory optimization problem in a static environment.
To demonstrate that our method is capable of obtaining a whole-body trajectory, using a mass point trajectory that even does not consider the robot's shape as the initial value.
The trajectory optimization process for this experiment is illustrated in the Fig.\ref{fig:car_optimization}(a)-\ref{fig:car_optimization}(f).
Based on a naive blue initial trajectory, as the number of iterations increases, 
the states on the trajectory that do not meet the scale constraint disappear.
As indicated in the Fig.\ref{fig:car_optimization}{(f)} and Fig.\ref{fig:car_optimization}{(g)}, the final optimization result is smooth and satisfies the whole-bdoy requirements.

\subsection{Experiment in Dynamic Environment}

\begin{figure}[t]
	\centering
    \vspace{0.2cm}
	\includegraphics[width=1\linewidth]{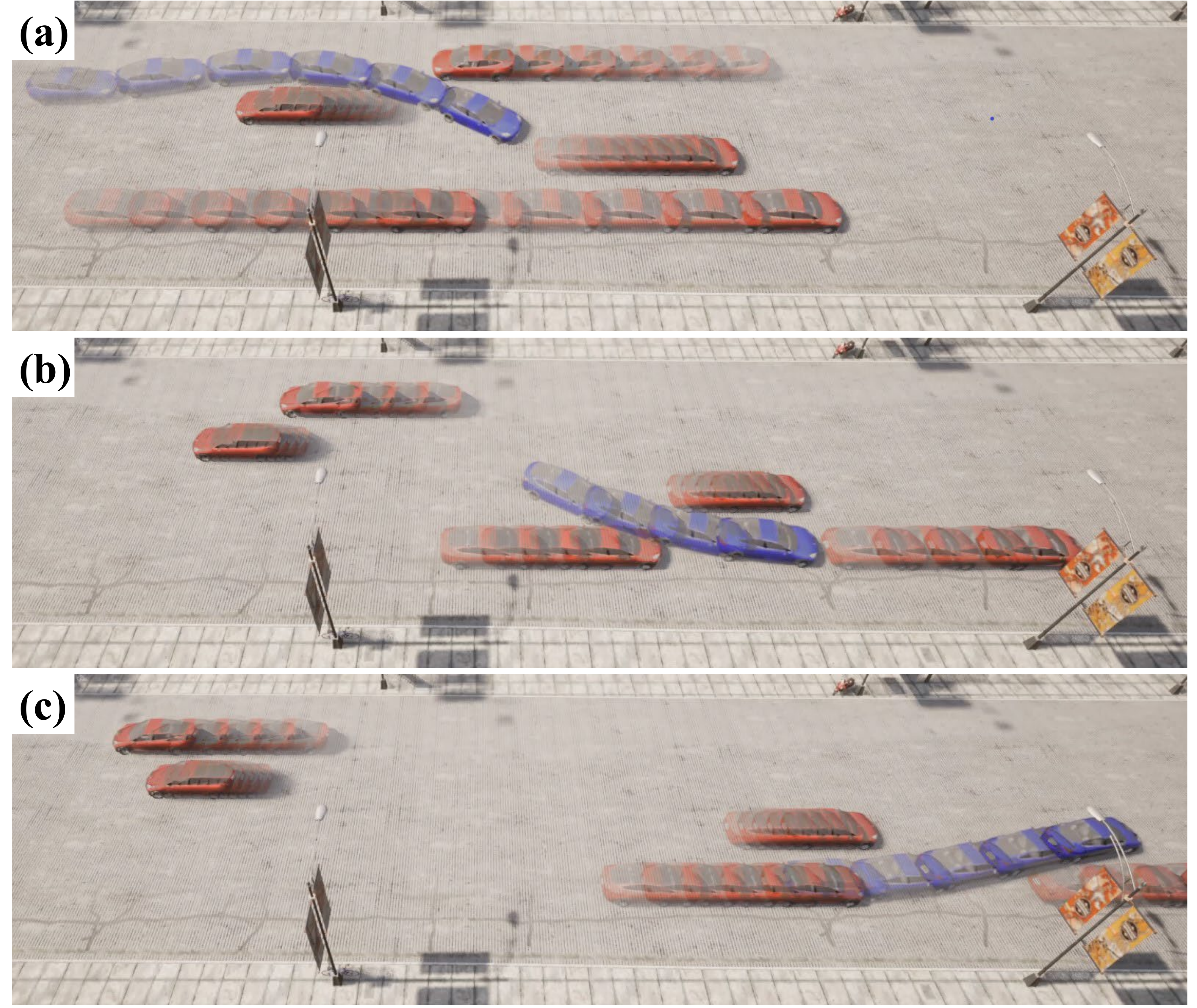}
    \vspace{-0.6cm}
	\caption{
        The snapshot of the experimental validation of our method applied in a vehicle robot.
        We plan the trajectory for the blue vehicle to navigate through the dynamic environment.
	}
	\label{fig:car}
    \vspace{-1cm}
\end{figure}

As illustrated in Fig.\ref{fig:car}, we employ our method to the blue vehicle's whole-body trajectory optimization problem, which requires the blue vehicle to traverse through the traffic flow consisting of five red vehicles with maximum speeds of $4, 1, 1.5, 4, 5.5~m/s$.
We set the maximum velocity $8m/s$ and acceleration $2m/s^2$ for the blue vehicle in optimization.
For a clear demonstration of the navigating process in the dynamic environment, as shown in Fig.\ref{fig:car}, we use three temporally consecutive images to show that the reslut where the blue vehicle's trajectory is safe and smooth.

\section{Conclusion and Future Work}
In this paper, we propose an exact whole-body collision formulation via linear scale, which can be solved efficiently.
Furthermore, we derive its analytic gradient and applied it to the trajectory optimization in aerial and vehicle robots.

In addition to the applications mentioned in Sec. \ref{sec:application} and Sec.~\ref{sec:application_car}, the proposed method can be applied to other kinds of robots, such as manipulators and legged robots.
Moreover, benefiting from its scale-based design, this method can be implemented for deformable robots and swarm formations as well.
The above applications of this method will be released in the near future.
It is also worth mentioning that we will consider continuous collision formulation in trajectory optimization to improve the completeness of planning.

\bibliographystyle{IEEEtran}
\bibliography{references}
\end{document}